\documentclass[10pt,twocolumn,letterpaper]{article}

\usepackage[final]{cvpr}      

\usepackage{booktabs}
\usepackage{siunitx}
\usepackage{threeparttable}
\usepackage{graphicx}
\usepackage{float}
\usepackage{capt-of}  
\usepackage{subcaption}
\usepackage{array}     
\usepackage{tabularx}  
\usepackage{colortbl}
\usepackage{booktabs}

\definecolor{cvprblue}{rgb}{0.21,0.49,0.74}
\usepackage[pagebackref,breaklinks,colorlinks,allcolors=cvprblue]{hyperref}

\title{A Physical Coherence Benchmark for Evaluating Video Generation Models via Optical Flow-guided Frame Prediction}

\author{
    Yongfan Chen\textsuperscript{1},
    Xiuwen Zhu\textsuperscript{2},
    Tianyu Li\textsuperscript{2},\\
    \textsuperscript{1}Zhejiang University, \quad
    \textsuperscript{2}Alibaba Group
}

\begin{document}

\twocolumn[{
    \renewcommand\twocolumn[1][]{#1}%
    \maketitle
    \vspace{-1pt}
    \begin{center}
        \centering
        \includegraphics[width=1.0\textwidth]{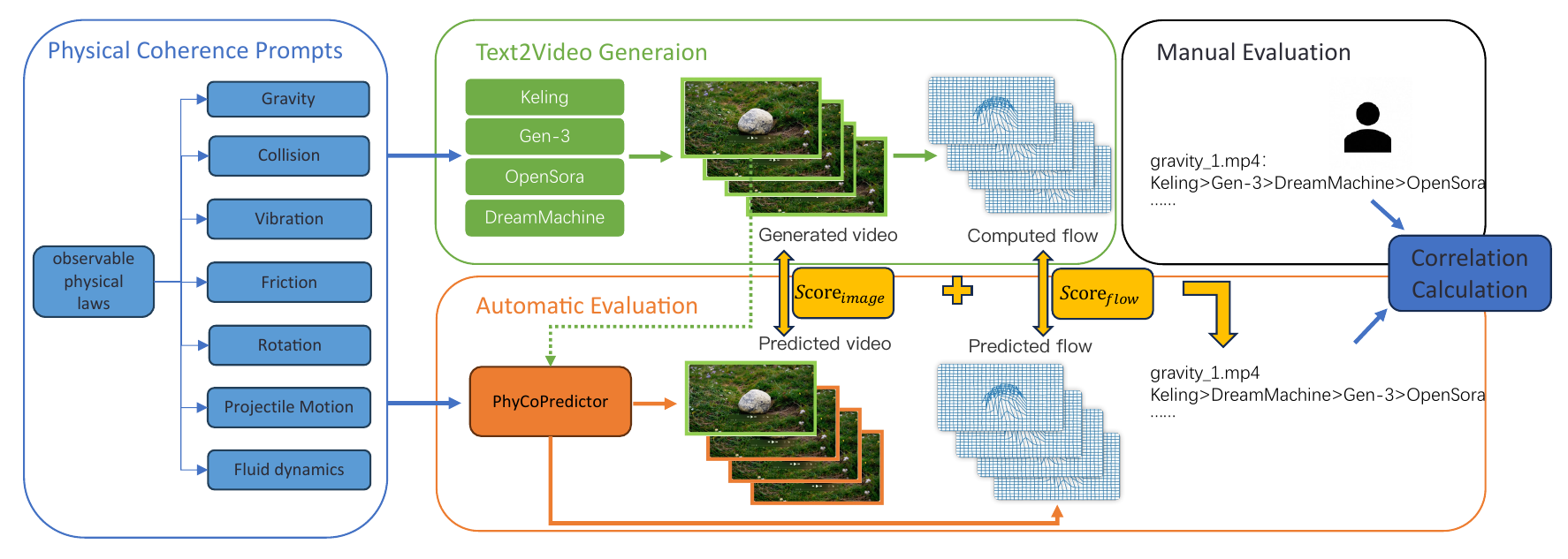}
        \vspace{-1em}
        \captionof{figure}{
            \textbf{Overview of Our Benchmark.} We propose \textbf{PhyCoBench}—a benchmark specifically designed to evaluate text-to-video (T2V) models in generating physically coherent videos. We categorize common physical scenarios into seven types and create \textbf{a comprehensive set of prompts}. With these prompts, we generate test set videos with four T2V models and conduct human rankings. We also introduce \textbf{PhyCoPredictor}, an optical flow-guided frame prediction model designed for automatic evaluation. Using the first frame and corresponding prompt of each video in the test set as input, we generate reference optical flow and videos with PhyCoPredictor. These are compared with the test set videos and their computed optical flows, producing scores to rank model performance. Correlation analysis shows that our automated evaluation results are \textbf{closely aligned with human preferences}.
        }
        \label{fig:teaser}
    \end{center}
    \vspace{-1pt}
}]

\begin{abstract}
Recent advances in video generation models demonstrate their potential as world simulators, but they often struggle with videos deviating from physical laws, a key concern overlooked by most text-to-video benchmarks. We introduce a benchmark designed specifically to assess the \textbf{Phy}sical \textbf{Co}herence of generated videos, \textbf{PhyCoBench}. Our benchmark includes 120 prompts covering 7 categories of physical principles, capturing key physical laws observable in video content. We evaluated four state-of-the-art (SoTA) T2V models on PhyCoBench and conducted manual assessments. Additionally, we propose an automated evaluation model: PhyCoPredictor, a diffusion model that generates optical flow and video frames in a cascade manner. Through a consistency evaluation comparing automated and manual sorting, the experimental results show that PhyCoPredictor currently aligns most closely with human evaluation. Therefore, it can effectively evaluate the physical coherence of videos, providing insights for future model optimization. Our benchmark, including physical coherence prompts, the automatic evaluation tool PhyCoPredictor, and the generated video dataset, has been released on GitHub at \href{https://github.com/Jeckinchen/PhyCoBench}{https://github.com/Jeckinchen/PhyCoBench}.

\end{abstract}
\section{Introduction}
With the rapid advancement of video generation models \cite{animatediff, dynamicrafter, text2video, tuneavideo, videocrafter1, videocrafter2}, the quality of generated videos has continuously improved, demonstrating their potential to become world simulators. However, ensuring the physical coherence of video content remains challenging, which is one of the major concerns for current video generation models. Physical coherence refers to the extent to which the motion in a video follows physical laws observed in real-world scenarios. However, most video generation benchmarks \cite{2024vbench, evalcrafter, t2vbench} do not evaluate physical coherence, allowing visually appealing but physically implausible content to receive high scores. Recently, VIDEOPHY \cite{videophy} attempted to address this issue using video-language models to answer "Does this video follow the physical laws?" The logits output can be used to assess the model's decision tendency, but this score is not equivalent to a metric that indicates how well a video conforms to physical laws.

To comprehensively evaluate whether generated videos adhere to physical laws, we introduce a novel video generation benchmark called PhyCoBench. PhyCoBench aims to encompass observable physical phenomena, including Newton's laws of motion, conservation principles (energy, momentum), collisions, rotational motion, static equilibrium, elasticity, vibration, and fluid dynamics. To capture these physical phenomena in specific human or object interactions, we construct a text prompt benchmark set across three categories: (1) simulated physical experiments, (2) common physical phenomena in everyday life, and (3) object movements in sports activities. The primary motion types examined in these test cases can be categorized into seven major groups: (1) gravity, (2) collision, (3) vibration, (4) friction, (5) fluid dynamics, (6) projectile motion, and (7) rotation.
We present PhyCoBench, a benchmark specifically designed to evaluate the physical consistency of generated videos. Our benchmark covers seven categories of physical principles: gravity, collision, vibration, friction, rotation, projectile motion, and fluid dynamics,capturing the majority of physical laws that are readily observable in video content. A total of 120 prompts are provided, each exemplifying the corresponding physical principles across these categories.

We create a benchmark set of 120 prompts for these seven categories. The content of these prompts draws inspiration from various motion recognition datasets, which are expanded using large language models (LLMs) \cite{qwen2} and further refined by human experts. Based on these prompts, we use state-of-the-art video generation models to produce corresponding videos. Our goal is to evaluate these models' ability to generate physically consistent content.

Furthermore, we propose an optical flow-guided video frame prediction model. In video generation tasks, the appearance and texture of objects can distract the model from focusing on physical laws, making it difficult for the generated video to follow these laws. To address this, we guide the generation process using optical flow, which contains only motion information. This allows the model to focus on object motion trajectories, thereby enhancing its adherence to physical laws.

Similar to video anomaly detection tasks, evaluating the physical coherence of generated videos requires detecting anomalies within the video. However, this task presents two main challenges: (1) anomalies are diverse and complex, making them difficult to define and quantify; (2) existing datasets lack negative samples, hindering the model's ability to learn prior knowledge of anomalies. To address these issues, we take inspiration from previous video anomaly detection approaches \cite{hf2vad, liu2018future, nguyen2019anomaly, zaheer2020old, yang2023video} and propose a frame prediction model called PhyCoPredictor to detect anomalies by predicting future frames. Specifically, our model takes an initial frame and a text prompt as inputs, utilizes a latent diffusion model to predict future optical flow, and uses the predicted flow to guide a text-conditioned video diffusion model for future frame prediction. After extensive training, our model can effectively predict the optical flow and visual content in dynamic scenes, allowing comparison with existing video generation models to evaluate physical coherence.

Overall, our contributions are as follows:
\begin{itemize}
    \item We construct \textbf{a comprehensive benchmark set of text prompts} for physical scenarios, which covers a wide range of common motion scenes.
    \item We propose an optical flow-guided model, named \textbf{PhyCoPredictor}, for video frame prediction that effectively predicts motion information in dynamic scenes.
    \item Using the proposed prompts and model, we develop a benchmark, named \textbf{PhyCoBench}, for evaluating the physical coherence capabilities of video generation models.
    \item Consistency evaluation shows that PhyCoPredictor can effectively assess  T2V model's ability to generate videos that satisfy physical laws.

\end{itemize}
\section{Related Work}
\label{sec:relatedwork}

\begin{figure}[t]
  \centering
    \vspace{-10pt}
   \includegraphics[width=0.6\linewidth]{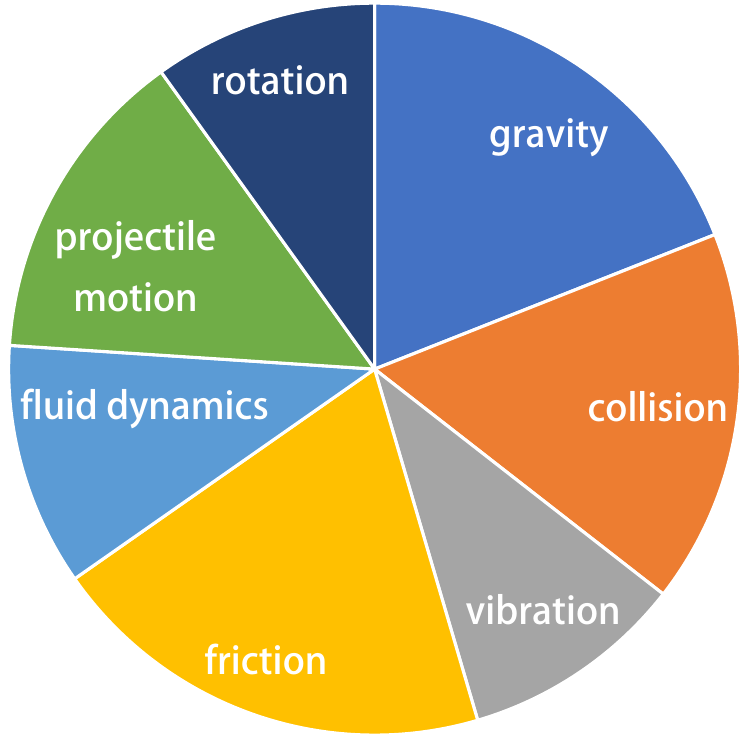}
   \vspace{-5pt}
   \caption{
   \textbf{The proportion of text prompts.} Our prompts are grouped into seven types.
   }
    \label{fig:prompt}
    \vspace{-10pt}
\end{figure}

\begin{table*}[hbt!]
\centering
\fontsize{6}{7}\selectfont  
\renewcommand{\arraystretch}{0.9}  
\resizebox{\textwidth}{!}{
\begin{tabular}{c >{\centering\arraybackslash}p{0.45\textwidth}}  
\hline
\textbf{Category} & \textbf{Example Prompts} \\
\hline
Gravity & A coin falls freely from a height and lands on the carpet. \\
       & A little girl throws a tennis ball into the air, and it falls back down. \\
\hline
Collision & A tennis ball hits the ground and bounces back up. \\
          & Two billiard balls collide on the table and separate. \\
\hline
Vibration & A swing starts to sway after being gently pushed. \\
          & A pendulum swings continuously from side to side. \\
\hline
Friction & A coin slides down a smooth, steep metal slide. \\
          & A stone slides down a rough, gentle dirt slope. \\
\hline
Fluid dynamics & A stone falls into a pool. \\
                & Milk is poured into coffee. \\
\hline
Projectile motion & A girl throws a basketball into the hoop. \\
                  & A car drives off a cliff. \\
\hline
Rotation & An airplane engine starts, and the blades begin to turn. \\
         & A merry-go-round spins at high speed. \\
\hline
\end{tabular}
}
\caption{Example Prompts of PhyCoBench for Different Categories}
\label{tab:example_prompts}
\end{table*}

\subsection{Reconstruction and Prediction-Based Video Anomaly Detection}

The fundamental principle of reconstruction and prediction-based video anomaly detection methods is to to train models on a large number of normal videos, enabling them to reconstruct or predict normal frames. When an abnormal frame is present in the input video, it can be detected by comparing it with the model's output. For example, based on the previous frame, \cite{liu2018future} predicts the next frame using FlowNet\cite{flownet} and GANs\cite{GANs}. \cite{nguyen2019anomaly} predicts both the next frame and the optical flow between consecutive frames. \cite{zaheer2020old} generates high-quality future frames using GANs. \cite{hf2vad} reconstructs optical flow using a memory module and employs a conditional VAE to predict future frames. \cite{yang2023video} selects multiple keyframes from $t$ frames to reconstruct the original sequence.

In this paper, we draw inspiration from the approaches mentioned above and use a frame prediction model to detect whether generated videos exhibit physical coherence. Instead, we employ a latent diffusion model-based approach, which makes our model more flexible and efficient.

\subsection{Video Generation Model}

The field of video generation has been developing rapidly \cite{VideoLDM, videocrafter1, videocrafter2, seer, aid, animatediff, harvey2022flexible, imagen, ho2022video, text2performer, text2video, singer2022make, 3modelscope, dynamicrafter, magicvideo, anonymous2024d}, with numerous studies on text-to-video\cite{text2video, tuneavideo, videocrafter1, videocrafter2} and image-to-video generation\cite{dynamicrafter, animatediff, chen2023motion}. VDM \cite{ho2022video} is the first to introduce a diffusion-based approach for video generation, laying the foundation for subsequent developments in this direction. Subsequently, works like MagicVideo \cite{magicvideo}, Video LDM \cite{VideoLDM}, and LVDM \cite{LVDM} introduce latent diffusion approaches that generate videos in latent space. SVD\cite{svd} standardize the pre-training process for video generation models in three steps: text-to-image pre-training, video pre-training on a large-scale low-resolution dataset, and fine-tuning on a small-scale high-quality high-resolution dataset. Sora \cite{sora} advance the diffusion Transformer models by replacing the U-Net architecture in the latent diffusion model (LDM) with a Transformer while directly compressing videos using a video encoder.

The rapid progress in video generation models has placed higher demands on video generation benchmarks. However, existing benchmarks still struggle to comprehensively evaluate model capabilities and the quality of generated videos.

\subsection{Automatic Metrics for Video Generation}

Current popular benchmarks primarily use the following metrics: IS \cite{IS}, FID \cite{FID}, FVD \cite{FVD}, SSIM \cite{SSIM}, CLIP-Score \cite{clipscore}, and their variants. EvalCrafter \cite{evalcrafter} includes 500 prompts, derived from both real user data and data generated with the assistance of large language models (LLMs), and uses 17 metrics to evaluate the capabilities of text-to-video generation models. Vbench \cite{2024vbench} assesses video generation quality from 16 dimensions, covering 24 categories with a total of 1,746 prompts. T2VBench \cite{t2vbench} contains 1,680 prompts to specifically evaluate text-to-video models in terms of temporal dynamics across 16 temporal dimensions. VIDEOPHY \cite{videophy} is a recently proposed text-to-video benchmark, consisting of 9,300 generated videos that are manually labeled. It evaluates video reasonableness based on human judgment of whether the videos conform to physical commonsense.

While these benchmarks evaluate video generation quality from multiple aspects, they still lack a simple and effective way to assess the physical coherence of video generation models. Therefore, we propose a new benchmark to fill this gap.

\begin{figure*}[ht]
\centering
\includegraphics[width=0.6\linewidth]{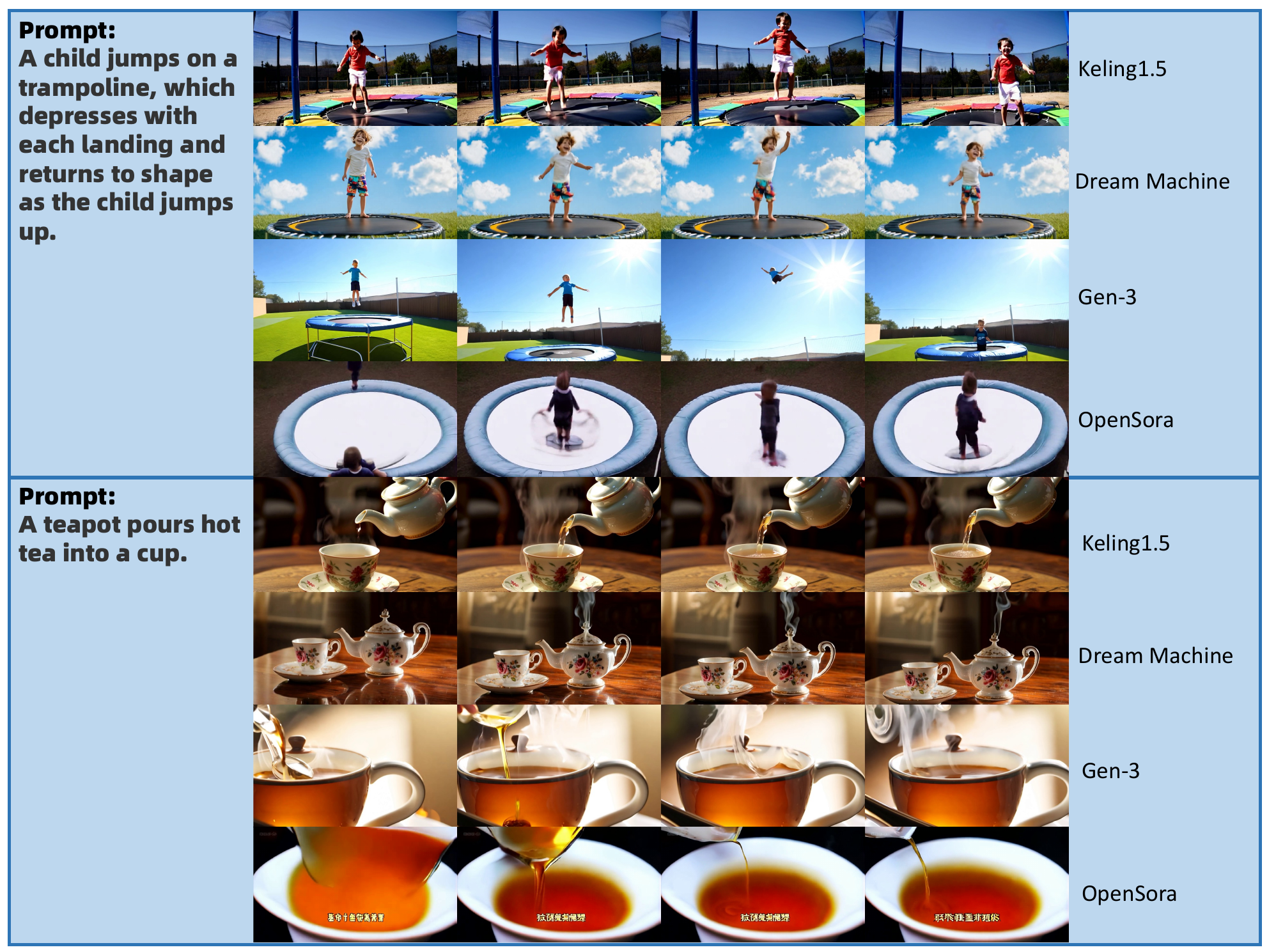}
\caption{\textbf{Generated video examples of T2V models.} The videos generated by these four models do not consistently adhere to physical coherence, with varying levels of quality.}
\label{fig:sota_vis}
\end{figure*}
\section{Physical Coherence Benchmark}

\subsection{Prompts}
To comprehensively evaluate the physical coherence of text-to-video generation models, we propose a benchmark set containing 120 prompts, categorized into seven groups: (1) gravity, (2) collision, (3) vibration, (4) friction, (5) fluid dynamics, (6) projectile motion, and (7) rotation. Some examples are shown in Table \ref{tab:example_prompts}. We reference definitions and explanations of motion from physics textbooks from a professional standpoint, while also considering common motion scenarios in action recognition datasets such as UCF101\cite{ucf101}, PennAction\cite{pennaction}, and HAA500\cite{haa500} from an everyday perspective. Ultimately, based on these references, we classify motions into seven categories. Our prompts can also be grouped into three types based on their content: (1) simulated physical experiments (e.g., "A rubber duck falls freely from a height and lands on the wooden floor."), (2) common physical phenomena in daily life (e.g., "A swing is pulled to the highest point and then released, beginning to sway."), and (3) object movements in sports activities (e.g., "A ping pong ball falls from a height onto a table and bounces."). The statistics for these prompts, in terms of their distribution across the seven categories, are shown in Figure~\ref{fig:prompt}.

\subsection{Human Evaluation Results}
We evaluated four text-to-video generation models (Keling1.5 \cite{keling}, Gen-3 Alpha \cite{runway}, Dream Machine \cite{luma}, and OpenSora-STDiT-v3 \cite{opensora}) by generating videos based on our 120 prompts. Some of the generated video results are shown in Figure~\ref{fig:sota_vis}. It is evident that the generated videos do not consistently adhere to physical consistency. For the same prompt, the quality of the videos varies significantly across the four models. This indicates that there is still a substantial gap between different models in terms of physical consistency. Therefore, there is an increased need for a more accurate and in-depth evaluation of model performance in this dimension.

\begin{figure}[ht]
\centering
\includegraphics[width=0.8\linewidth]{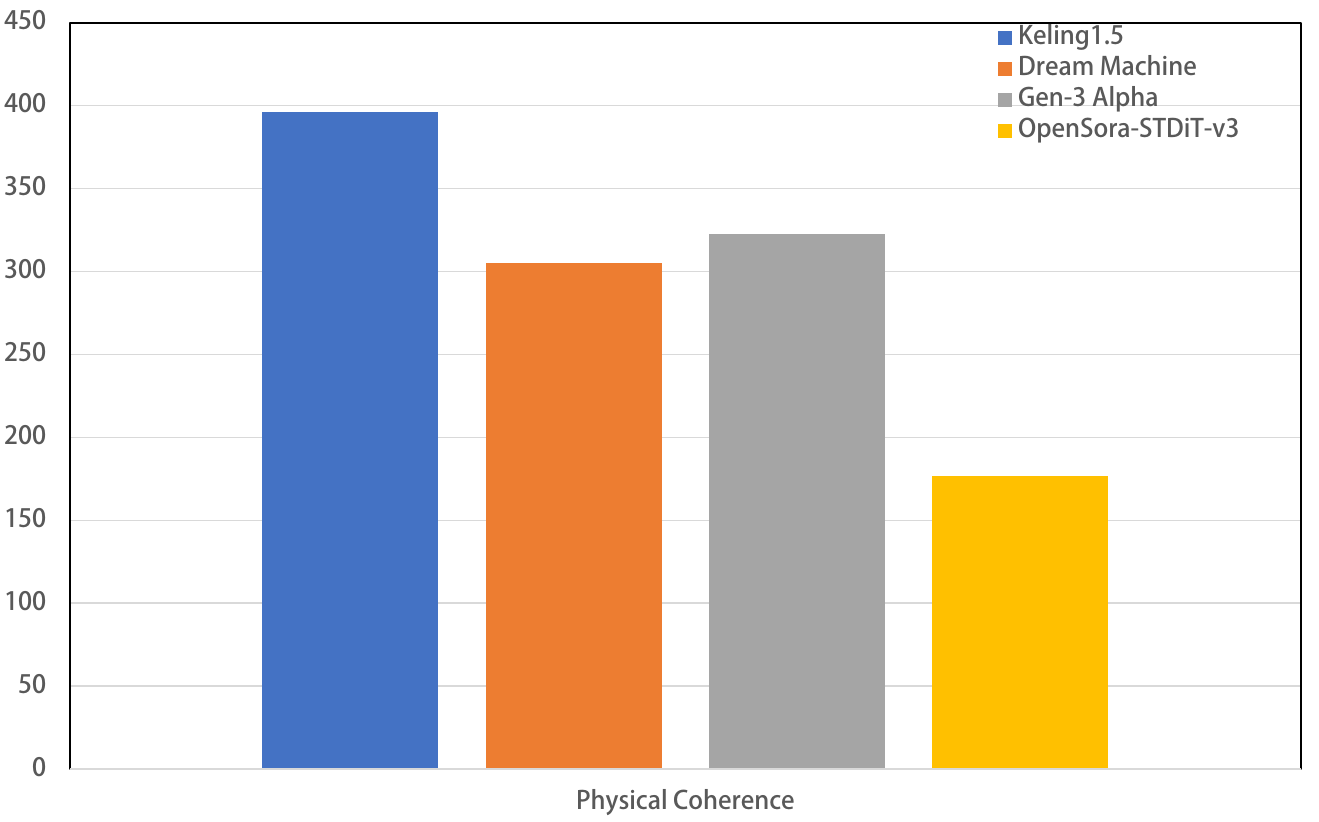}
\caption{\textbf{Overall ranking result from manual evaluation.}}
\label{fig:rank_all}
\end{figure}

\begin{figure}[ht]
\centering
\includegraphics[width=0.8\linewidth]{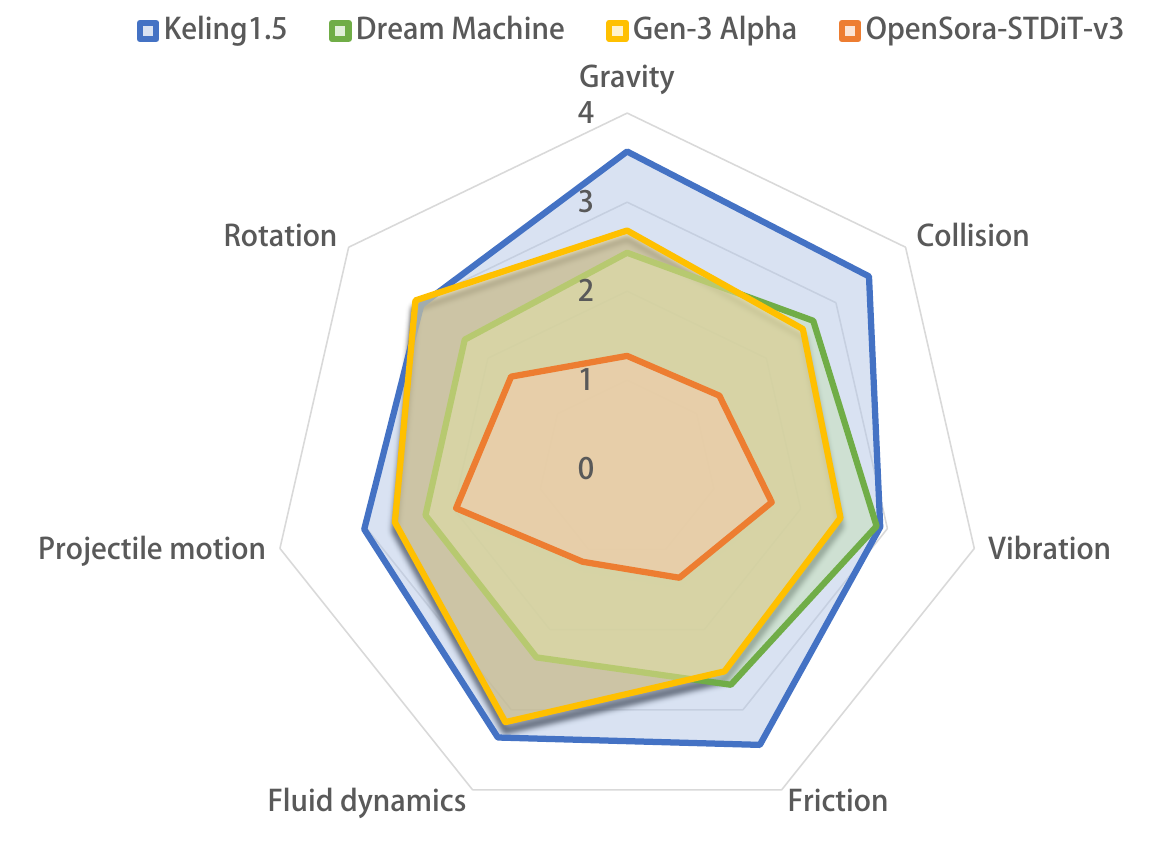}
\caption{\textbf{Category-specific ranking results from manual evaluation.}}
\label{fig:rank_c}
\end{figure}

For the videos generated by the four T2V models, we initially conduct a manual ranking of the four models for each prompt. The results are shown in Figure~\ref{fig:rank_all} and Figure~\ref{fig:rank_c}. 

In Figure~\ref{fig:rank_all}, the physical coherence performance of the four T2V models was manually evaluated and ranked. As shown in the figure, Keling1.5 stands out with the highest physical coherence, significantly outperforming the other models. The performances of Dream Machine and Gen-3 Alpha are quite close to each other. In contrast, the open-source model OpenSora-STDiT-v3 scores relatively lower, far behind the top three, indicating substantial room for improvement in terms of physical coherence.

Figure~\ref{fig:rank_c} presents the detailed performance of each model across the seven physical scenario categories. In this radar chart, Keling1.5 demonstrates the most comprehensive performance, covering the widest range, and achieves the highest scores in several categories, particularly excelling in gravity, collision, and friction. Dream Machine and Gen-3 Alpha show relatively balanced performance but are slightly behind. OpenSora-STDiT-v3, on the other hand, performs relatively poorly, failing to achieve high scores across all categories.

The drawback of manual evaluation is the lack of quantifiable metrics for comparison, as well as the high cost. Therefore, we use a frame prediction model for automated quantitative evaluation. Next, we will provide a detailed introduction to the frame prediction model.

\begin{figure*}[t]
  \centering
  \vspace{-1pt}
   \includegraphics[width=0.99\linewidth]{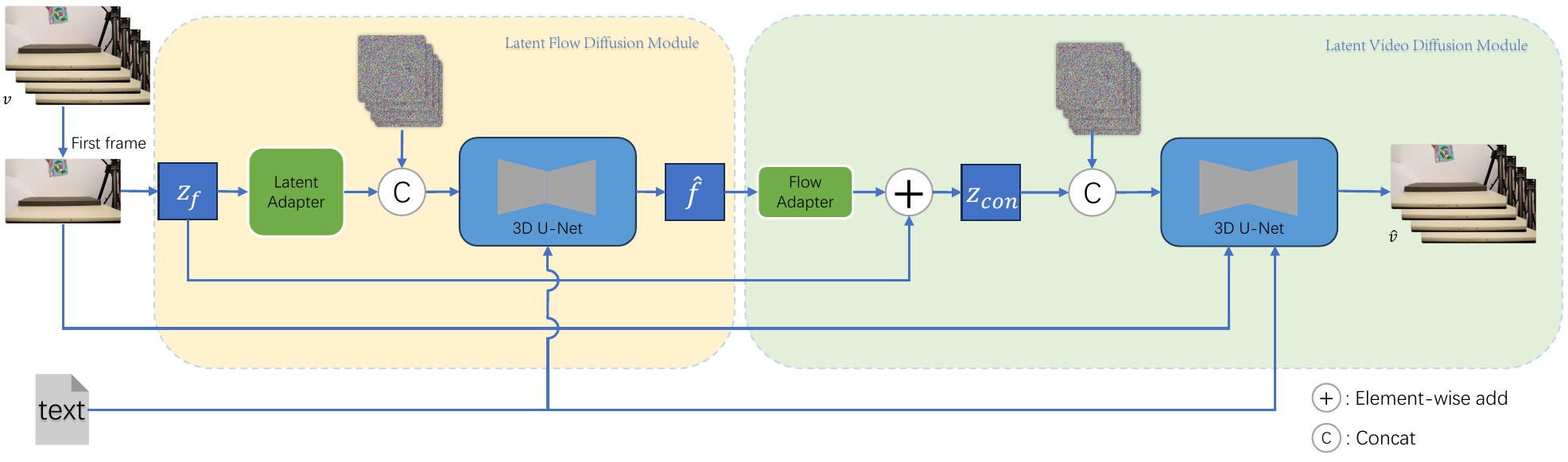}
   \vspace{-5pt}
   \caption{
   \textbf{Inference process of PhyCoPredictor.} Once we obtain the generated video from the T2V model, we input the first frame and the prompt into PhyCoPredictor. The Latent Flow Diffusion Module predicts the future optical flow, which then guides the Latent Video Diffusion Module to predict future video frames.}
   \label{fig:infer_pipe}
   \vspace{-1pt}
\end{figure*}
\section{Automatic Evaluator}

\begin{figure*}[t]
  \centering
  \vspace{-10pt}
   \includegraphics[width=0.99\linewidth]{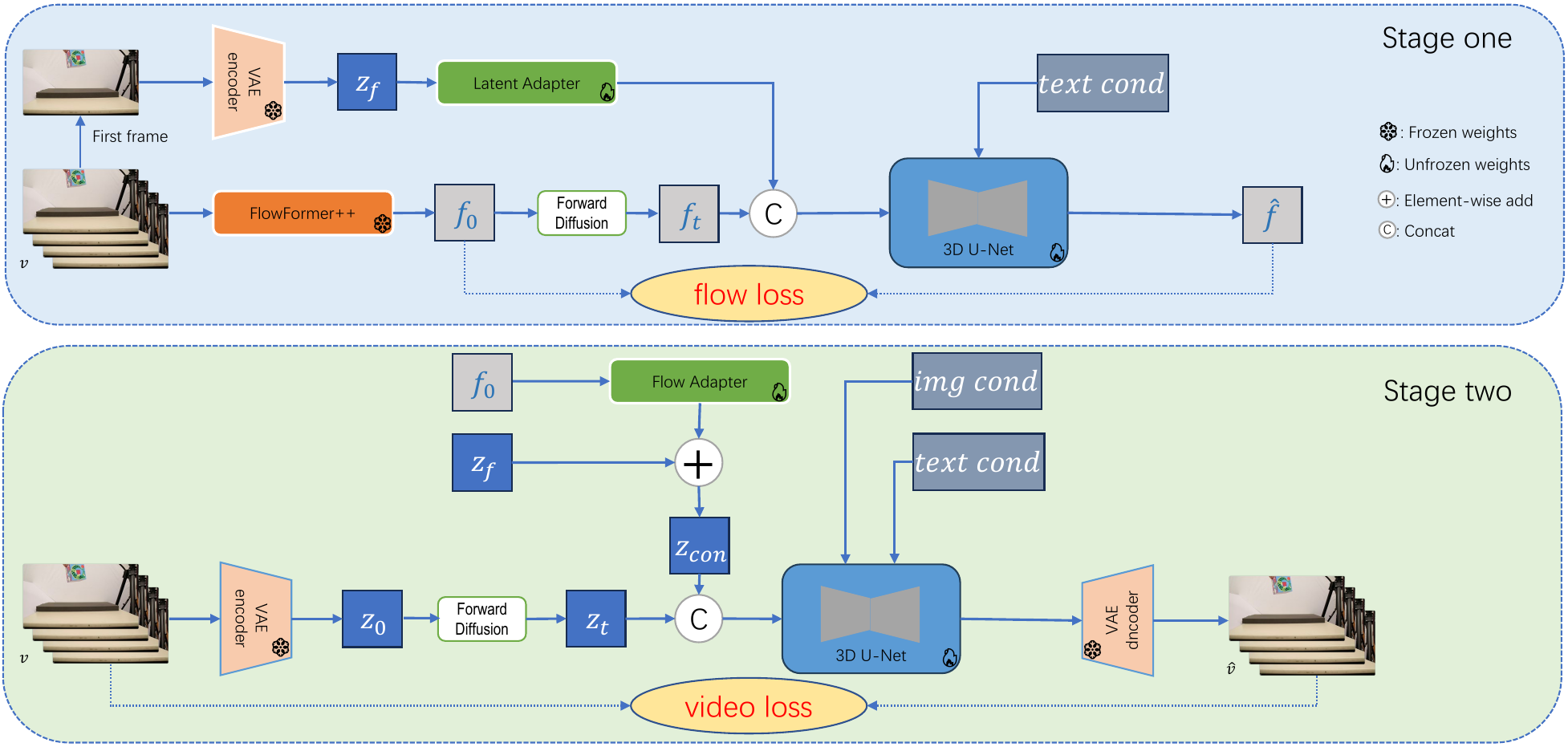}
   \vspace{-5pt}
   \caption{
   \textbf{Training pipeline.} Our model training is divided into two stages. In the first stage, we train the 3D U-Net from scratch to predict future optical flow. In the second stage, we use the pre-trained weights from DynamiCrafter and train the model to generate future video frames with more natural motion trajectories under the guidance of optical flow.}
   \label{fig:train_pipe}
   \vspace{-10pt}
\end{figure*}

Our optical flow-guided video frame prediction model, named PhyCoPredictor, comprises two Latent Diffusion Model (LDM) modules. First, after obtaining the video generated by a text-to-video model, we take the first frame of the input video along with the corresponding text prompt. The first LDM is used to predict the future optical flow from this initial frame. Then, the predicted future optical flow, combined with the initial frame and text prompt, serves as a guiding condition for the second LDM, which generates a multi-frame video starting from that frame. The model's training pipeline is shown in Figure~\ref{fig:train_pipe}. Our ultimate goal is to determine whether the generated video maintains physical coherence by detecting physical inconsistencies or anomalies. To achieve this, we use FlowFormer++ \cite{shi2023flowformer++} to compute the optical flow of the generated video as a reference flow, and compare the reference flow, the generated video, the predicted flow, and the predicted video. If the discrepancies are significant, it indicates that the generated video contains anomalies and does not meet physical coherence.

\subsection{Preliminary: Latent Diffusion Model}

Our model belongs to the class of generative diffusion models \cite{ho2020denoising}. Diffusion models define both a forward diffusion process and a reverse denoising process. The forward diffusion process gradually adds noise to the data \( x_0 \sim p(x) \), which resulting in Gaussian noise \( x_T \sim \mathcal{N}(0, I) \), while the reverse denoising process restores the original data by progressively removing noise. The forward process \( q(x_t \mid x_0, t) \) consists of \( T \) timesteps, during which at each timestep, \( x_{t-1} \) is gradually noised to obtain \( x_t \). The denoising process \( p_\theta(x_{t-1} \mid x_t, t) \) uses a denoising network \( \epsilon_\theta(x_t, t) \) to predict a less noisy version \( x_{t-1} \) from the noisy input \( x_t \). The objective function of the denoising network is 
\begin{equation}
    \min_{\theta} \mathbb{E}_{t, x \sim p, \epsilon \sim \mathcal{N}(0, I)} \left\| \epsilon - \epsilon_\theta(x_t, t) \right\|_2^2,
\end{equation}
where \( \epsilon \) represents the true noise, and \( \theta \) is the set of learnable parameters of the network. After training, the model can employ the reverse denoising process to recover the original data \( x_0 \) from random Gaussian noise \( x_T \).

To reduce computational complexity, the Latent Diffusion Model (LDM) \cite{LDM} was proposed, which performs noise addition and denoising in the latent space. Many recent works on diffusion models are based on the LDM architecture, including our work. In this paper, our frame prediction model is built upon an open-source image-to-video LDM framework called DynamiCrafter \cite{dynamicrafter}.  For LDM, the input \( x_0 \) is encoded to obtain the latent variable \( z_0 = E(x) \). The forward noise addition process \( q(z_t \mid z_0, t) \) and the reverse denoising process \( p_\theta(z_{t-1} \mid z_t, t) \) are performed in the latent space, and the final output of the model is obtained by decoding with a decoder, \( \hat{x} = D(z) \).

\subsection{PhyCoPredictor}

\subsubsection{Latent Flow Diffusion Module}

To enhance the performance of the frame prediction model, we drew inspiration from 
 \cite{lfdm} by using optical flow as a condition to guide the frame prediction process. The optical flow mentioned here is the latent flow generated by LDM. We input the text prompt and the first frame \( x \in \mathbb{R}^{1 \times 3 \times H \times W} \) of the video into the model, and after encoding through the VAE encoder, we replicate it \( N \) times to obtain the latent variable \( z_f \in \mathbb{R}^{N \times 4 \times h \times w} \), which contains the visual information of the first frame. Next, we compute the optical flow for \( N \) frames using FlowFormer++ and downsample it to the latent space, yielding the flow \( f \in \mathbb{R}^{N \times 2 \times h \times w} \), where \( f_0 \) is the optical flow calculated between the first frame and itself, and \( f_i \) (\( i \neq 0 \)) represents the flow calculated between the \((i - 1)\)-th and \( i \)-th frames.

To align the dimensions of the latent variable and the optical flow, we designed a Latent Adapter consisting of an MLP layer and an activation layer, which downsamples the feature dimension of \( z_f \) from 2 to 4. Next, the noised latent flow \( f \) is concatenated with \( z_f \), and the result is fed into a 3D U-Net. After the denoising process, we obtain the generated optical flow \( \hat{f} \in \mathbb{R}^{N \times 2 \times h \times w} \). Our loss function is 
\begin{equation}
\mathcal{L}_\text{flow} = \| f - \hat{f} \|_2^2,
\end{equation}
where \( \hat{f} \) represents the latent flow predicted by the model.

\subsubsection{Latent Video Diffusion Module}

Our network architecture for predicting video frames in latent space is built upon DynamiCrafter \cite{dynamicrafter}. To more accurately predict the motion trajectories of objects in physically dynamic scenes, we chose to use optical flow as a condition to guide the frame prediction process. During the training phase, we utilize FlowFormer++ \cite{shi2023flowformer++} to obtain optical flow and employ a Flow Adapter to upsample the feature dimension from 2 to 4 to align with the latent space. Our Flow Adapter consists of a 3D convolution layer. We then add the upsampled flow to the latent variable \( z_f \), resulting in \( z_{\text{con}} \in \mathbb{R}^{N \times 4 \times h \times w} \), which fuses both motion and visual information. The input consists of a text prompt and video \( v \in \mathbb{R}^{N \times 3 \times H \times W} \). The video \( v \) is encoded by a VAE encoder to produce \( z_0 \in \mathbb{R}^{N \times 4 \times h \times w} \). After adding noise to \( z_0 \), it is concatenated with \(  z_{\text{con}} \) and fed into a 3D U-Net. Similar to DynamiCrafter, we use a CLIP image encoder to encode the first frame of the video, convert it into visual conditions through an image context network, and control the generation result through cross-attention alongside the text conditions provided by the text prompt. During this process, our loss function is
\begin{equation}
\mathcal{L}_\text{video} = \| v - \hat{v} \|_2^2,
\end{equation}
where \( \hat{v} \) represents the video predicted by the model.

During the inference phase, the optical flow is provided by the previously mentioned Latent Flow Diffusion Module (and needs to be upsampled from 32×32 to 40×64) to predict the motion trends of future video frames.

\subsubsection{Training Setup}

In our model, apart from the two 3D U-Nets and two adapters, all other components are frozen. The training process is divided into two stages: In the first stage, we train the Latent Adapter and the 3D U-Net in the Latent Flow Diffusion Module. We initially use LLM\cite{qwen2} to filter out relatively static data based on the captions from Openvid\cite{openvid}, and then train these two components from scratch using the remaining data. We sample 16 frames from the videos to ensure that the model focuses on dynamic content. The batch size is set to 4, and the training runs for a total of 100k steps on 64 L20 GPUs. Subsequently, we use Motion Data for an additional 30k steps of training, still with a batch size of 4 and on 64 L20 GPUs. The Motion Data is a dynamic scene dataset selected from UCF101\cite{ucf101}, Physics101\cite{phys101}, Penn Action\cite{pennaction}, and HAA500\cite{haa500}, annotated with captions using Multimodal Language Model(MLM). All training in the first stage is conducted at a resolution of 256×256, with the corresponding latent space dimensions being 32×32.

In the second stage, we train the Flow Adapter and the 3D U-Net in the Latent Video Diffusion Module. The U-Net in this stage is initialized with pre-trained weights from the DynamiCrafter model at a resolution of 320×512, with the latent space dimensions being 40×64. The training also uses Motion Data, with a batch size of 4 for a total of 30k steps.

\subsection{Automatic Evaluation Process}

We used FlowFormer++\cite{shi2023flowformer++} to compute the optical flow of the generated videos and sampled them to \( N \) frames to obtain the original optical flow, while also sampling the generated videos to \( N \) frames to obtain the original video. Subsequently, we input the first frame of the original video and the corresponding prompt into the trained frame prediction model to obtain the predicted optical flow and predicted video. To predict optical flow and video frames, we propose an optical flow-guided frame prediction model called PhyCoPredictor. The inference process of PhyCoPredictor is illustrated in Figure~\ref{fig:infer_pipe}.

Our model is trained extensively on typical motion scenarios to learn motion and visual priors. As a result, it generates predictions that are more likely to exhibit physically consistent behavior when forecasting optical flow and future frames. In cases where the original video does not satisfy physical coherence, anomalies can be detected by comparing the original optical flow and video with the predicted optical flow and video. We evaluate the performance of the four models by calculating both optical flow loss and video loss. Finally, we evaluate the performance of each model using a scoring metric defined as:
\begin{equation} 
\text{score} = \frac{1}{\text{MSE}(f, \hat{f})} + 2 \times \text{MSE}(v, \hat{v}), 
\end{equation}
where \( f \) and \( \hat{f} \) denote the original and predicted optical flow, respectively, while \( v \) and \( \hat{v} \) represent the original and predicted video frames, respectively. A higher score indicates better physical coherence of the generated video.

\subsection{Evaluation Results}

\subsubsection{Quantitative Evaluation}

Based on our proposed text prompt benchmark set, we evaluated the text-to-video generation performance of four models: Keling1.5, Gen-3 Alpha, Dream Machine, and OpenSora-STDiT-v3. For each model, we generated a video using each prompt. We then manually ranked the performance of the four models for each prompt. Additionally, we used Dynamicrafter and our proposed optical flow-guided frame prediction model to score and rank the four models. We calculated Kendall's Tau-b coefficient and Spearman's Rank Correlation coefficient to compare the model ranking results with the manual evaluations. The results in Table~\ref{tab:correlation_coefficients} indicate that our model's rankings align more closely with human assessments, and that incorporating optical flow guidance provides a better evaluation of the physical coherence of generated videos.

\begin{table}[ht]
\centering
\begin{threeparttable}
\begin{tabular}{l
                S[table-format=1.2, detect-weight]
                S[table-format=1.2, detect-weight]}
\toprule
\textbf{Method} & {\textbf{Kendall's} ($\uparrow$)} & {\textbf{Spearman's} ($\uparrow$)} \\
\midrule
\text{DynamiCrafter}$^*$ & \text{-0.2438} & \text{-0.2783} \\
\text{VideoPhy} & \text{0.0147} & \text{0.0429} \\
\textbf{Our Method} & \textbf{0.3367} & \textbf{0.3751} \\
\bottomrule
\end{tabular}
\begin{tablenotes}
\footnotesize
\item[*] indicates finetuned with the same training data.
\end{tablenotes}
\end{threeparttable}
\caption{Comparison of Model Rankings with Kendall's Tau-b and Spearman's Rank Correlation Coefficients}
\label{tab:correlation_coefficients}
\end{table}

\begin{figure*}[t]
  \centering
    \vspace{-10pt}
   \includegraphics[width=0.9\linewidth]{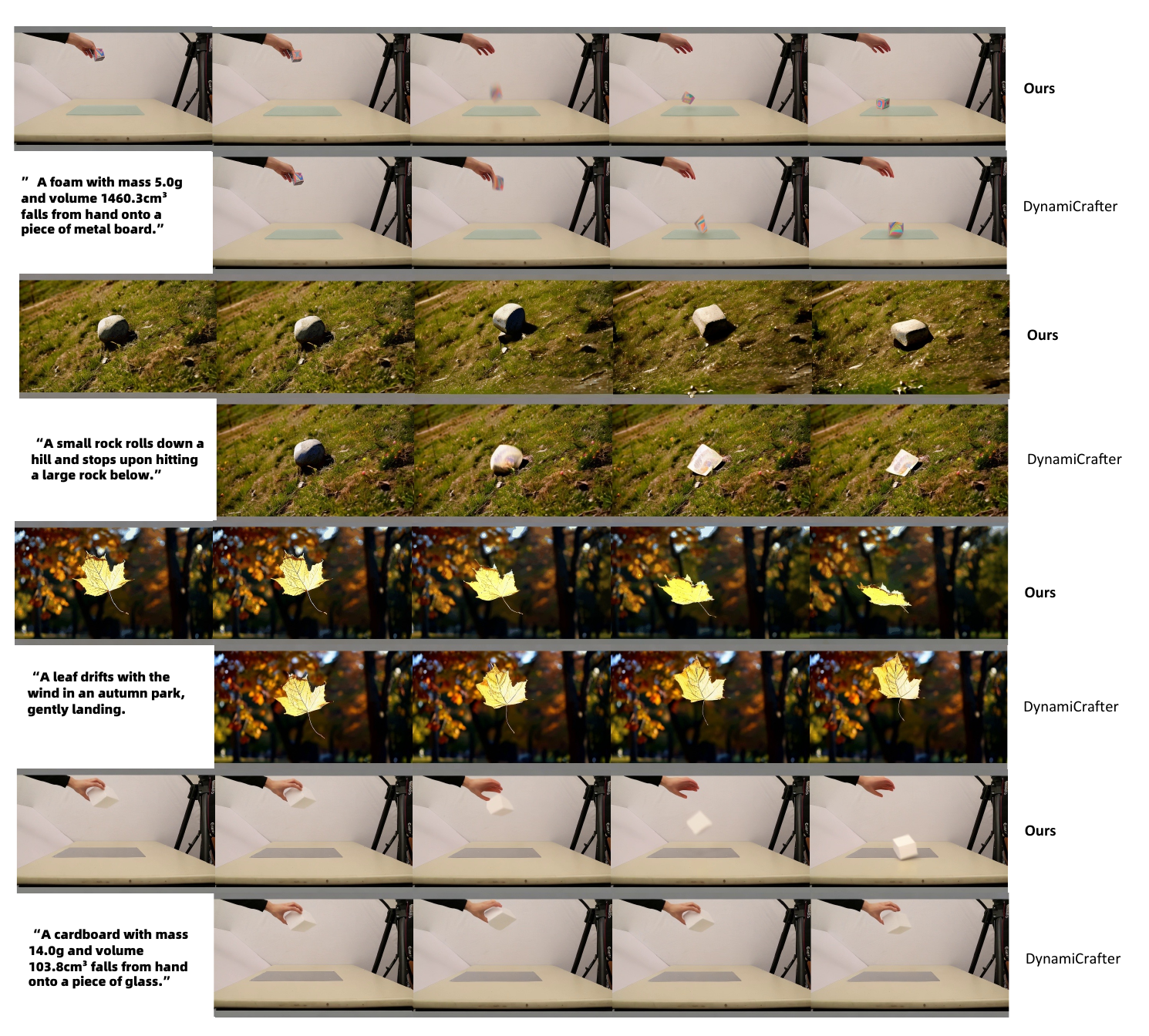}
   \vspace{-5pt}
   \caption{
   \textbf{Visual comparisons of frame prediction results from DynamiCrafter and our PhyCoPredictor.}
   }
    \label{fig:vis}
    \vspace{-10pt}
\end{figure*}

\subsubsection{Qualitative Evaluation}

To further demonstrate the effectiveness of our model, we conducted a visual comparison of video predictions using DynamiCrafter and our optical flow-guided frame prediction model. The visualization results in Figure~\ref{fig:vis} show that, with the addition of optical flow guidance, PhyCoPredictor produces more natural and realistic motion trajectories in the predicted video frames. For example, in scenes involving falling objects, it can accurately predict the falling trajectory and even the rebound trajectory after the fall. In contrast, DynamiCrafter struggles to effectively predict the motion trajectories of objects. In the scene depicting leaves falling naturally, PhyCoPredictor can accurately predict the leaves drifting down and swaying in the wind, while DynamiCrafter incorrectly predicts the leaves moving upward. Additionally, PhyCoPredictor can capture the complex motion of a rolling stone, whereas DynamiCrafter performs poorly in this scenario.

\section{Conclusion}

The rapid development of the video generation field has raised higher demands for video benchmarks, necessitating more comprehensive and effective methods for evaluating video quality. In this paper, we introduce PhyCoBench, a benchmark specifically focused on the dimension of physical coherence in videos. We have carefully designed a prompt set that comprehensively covers various physical scenarios. We also utilize an optical flow-guided frame prediction model, PhyCoPredictor, which effectively predicts the motion trajectories of objects. Based on this model and the prompt set, we can effectively assess whether the content of a video adheres to physical coherence. We believe that PhyCoBench makes a significant contribution to the fields of video generation and evaluation, and will aid in enhancing the capabilities of future video generation models.

{
    \small
    \bibliographystyle{ieeenat_fullname}
    \bibliography{main}
}

\clearpage
\setcounter{page}{1}
\maketitlesupplementary

\appendix

\definecolor{mycolor}{RGB}{220, 230, 241}  

In this supplementary material, Section~\ref{sec:data} provides a detailed explanation of how the dataset used to train our PhyCoPredictor was obtained and processed. In Section~\ref{sec:eval}, we elaborate on the details of model evaluation. Section~\ref{sec:training} presents the detailed training settings.

\section{Training Data}
\label{sec:data}

\subsection{OpenVid}

OpenVid \cite{openvid} is a large-scale video-text paired dataset, containing up to 10 million videos. We use OpenVid's data to train the 3D U-Net in the Latent Flow Diffusion Module from scratch. Based on our observations, although the quality of both the videos and captions in this dataset is high, most of the video content tends to be relatively static. However, our work focuses on evaluating the physical coherence of videos, specifically targeting dynamic scenarios. Therefore, to improve the training effectiveness of the frame prediction model, we filter the dataset to retain only those videos depicting relatively dynamic scenes. We provide the video captions from OpenVid to Qwen2 \cite{qwen2} for analysis to determine whether the caption describes a dynamic scene, using a prompt template shown in the Table \ref{tab:template_qwen} below.

\begin{table}[ht]
\centering
\setlength{\tabcolsep}{0pt}  
\fontsize{8}{10}\selectfont
\renewcommand{\arraystretch}{1.0}
\begin{tabularx}{\columnwidth}{@{}X@{}}  
\toprule
\rowcolor{mycolor}
\sloppy
The following is a description of the main content of the video. Please analyze the description and determine whether the video is dynamic or static based on significant changes in the movement trajectories of people or objects. For example, dynamic videos include noticeable actions such as objects colliding, falling, being thrown, vibrating, people running, fast-moving trains, or other significant movements. Static videos have little to no movement, such as someone using a phone or having a conversation. If judged as dynamic, return `The video is dynamic`; otherwise, return `The video is static`. Video caption: \textbf{caption} \\
\bottomrule
\end{tabularx}
\caption{\textbf{The prompt template for Qwen2 to determine the dynamics of a video.}}
\label{tab:template_qwen}
\end{table}

We filter out dynamic scene videos based on Qwen's responses. OpenVid originally contains 10 million data points, which are reduced to 1.2 million after filtering.

\subsection{Motion Data}

Another dataset we use, called Motion Data, is composed of UCF101 \cite{ucf101}, PennAction \cite{pennaction}, HAA500 \cite{haa500}, and  Physics101 \cite{pennaction}. The first three datasets are action recognition datasets, with each video labeled with an action category.To obtain video captions, we first input the video into a Multimodal Language Model (MLM) to generate a description, which is then fed into a Large Language Model (LLM) for further refinement. The prompt we use is shown in Table~\ref{tab:caption_prompt}. For these three datasets, we filter the videos based on their action categories, and the filtered action categories for each dataset are shown in Table \ref{tab:three_dataset}.

\begin{table}[ht]
\centering
\setlength{\tabcolsep}{0pt}  
\fontsize{8}{10}\selectfont
\renewcommand{\arraystretch}{1.0}
\begin{tabularx}{\columnwidth}{@{}X@{}}  
\toprule
\rowcolor{mycolor}
\sloppy
The prompt input to the MLM:\par
Provide a detailed description of the video in English, focusing on the movement information of the objects in the video.\par
The prompt input to the LLM:\par
\textbf{``The output of the MLM''} The text above is a description of a motion video. Please modify the description to meet the following requirements: 1. Retain information about the movement of objects and people. 2. Remove descriptions of people's clothing. 3. Remove descriptions of the environment and background.\par
\\ 
\bottomrule
\end{tabularx}
\caption{\textbf{The prompt for generating video captions.}}
\label{tab:caption_prompt}
\end{table}

\begin{table}[ht]
\centering
\setlength{\tabcolsep}{0pt}  
\fontsize{6}{8}\selectfont  
\renewcommand{\arraystretch}{1.05}  
\begin{tabularx}{\columnwidth}{@{}>{\centering\arraybackslash}p{0.3\columnwidth}@{}X@{}}
\toprule
\textbf{Dataset} & \textbf{Filtered Action Categories} \\
\midrule
UCF101 & Archery, Baseball Pitch, Basketball, Bench Press, Billiards, Bowling, Boxing Punching Bag, Boxing Speed Bag, Clean And Jerk, Field Hockey Penalty, Frisbee Catch, Golf Swing, Hammer Throw, Hammering, Hula Hoop, Javelin Throw, Juggling Balls, Nunchucks, Pizza Tossing, Shotput, Soccer Juggling, Soccer Penalty, Table Tennis Shot, Tennis Swing, Swing, Throw Discus, Trampoline Jumping, Volleyball Spiking, Yo-Yo \\
\midrule
PennAction & baseball\_pitch, baseball\_swing, bench\_press, bowl, clean\_and\_jerk, golf\_swing, jump\_rope, jumping\_jacks, pullup, pushup, situp, squat, strum\_guitar, tennis\_forehand, tennis\_serve \\
\midrule
HAA500 & ALS Ice Bucket Challenge, add new car tire, atlatl throw, axe throwing, badminton serve, badminton underswing, base jumping, baseball bunt, baseball catch catcher, baseball catch flyball, baseball catch groundball, baseball pitch, baseball swing, basketball dribble, basketball dunk, basketball hookshot, basketball jabstep, basketball layup, basketball pass, basketball shoot, beer pong throw, bike fall, billiard hit, BMX riding, bowling, bowls throw, card throw, cast net, closing door, cross country ski slide, cross country ski walk, curling follow, curling push, curling sweep, dart throw, dice shuffle reveal, dice stack shuffle, discuss throw, flipping bottle, flying kite, football catch, football throw, frisbee catch, frisbee throw, golf part, golf swing, guitar flip, guitar smashing, gym lift, gym pull, gym push, hammer throw, hammering nail, hanging clothes, high jump jump, high jump run, hopscotch skip, hopscotch spin, horizontal bar flip, horizontal bar jump, horizontal bar land, horizontal bar spin, hurdle jump, javelin throw, juggling balls, kick Jianzi, pancake flip, pétanque throw, pizza dough toss, play yo-yo, punching sandbag, punching speedbag, push wheelchair, push wheelchair alone, shotput throw, sledgehammer strike down, sling, slingshot, soccer dribble, soccer header, soccer save, soccer shoot, soccer throw, softball pitch, speed stack, squash backhand, squash forehand, swinging axe on a tree, tennis backhand, tennis forehand, tennis serve, throw boomerang, throw paper-plane, throwing bouquet, tire pull, tire sled, trampoline, volleyball overhand, volleyball pass, volleyball set, volleyball underhand, weightlifting hang, weightlifting overhead, weightlifting stand \\
\bottomrule
\end{tabularx}
\caption{\textbf{The filtered action categories in UCF101, PennAction, and HAA500.}}
\label{tab:three_dataset}
\end{table}

As for Physics101, it is a video dataset focused on physical experiment scenarios. We use all of the videos in this dataset. The dataset labels include object type, weight, material, and experiment category (such as falling or collision). Using these three labels, we manually annotate a caption for each video.

In summary, the number of videos used from each of the four datasets is shown in Table~\ref{tab:video_num}.

\begin{table}[ht]
\centering
\setlength{\tabcolsep}{0pt}  
\fontsize{8}{10}\selectfont  
\renewcommand{\arraystretch}{1.05}  
\begin{tabular}{@{}>{\centering\arraybackslash}p{0.4\columnwidth}@{}>{\centering\arraybackslash}p{0.4\columnwidth}@{}}
\toprule
\textbf{Dataset} & \textbf{Filtered video nums} \\
\midrule
UCF101 & 3950 \\
\midrule
PennAction & 954 \\
\midrule
HAA500 & 2080 \\
\midrule
Physics101 & 6075 \\
\bottomrule
\end{tabular}
\caption{\textbf{The number of videos used in the four datasets.}}
\label{tab:video_num}
\end{table}

\section{Evaluation Details}
\label{sec:eval}

\subsection{Manual Evaluation}
We input the carefully designed set of 120 prompts into four models—Keling1.5 \cite{keling}, Dream Machine \cite{luma}, Gen-3 Alpha \cite{runway}, and OpenSora \cite{opensora}. The generated videos are then provided to professional evaluators for manual ranking. For each set of four videos generated from a prompt, the instructions given to the evaluators are shown in Table ~\ref{tab:manual}.

\begin{table}[ht]
\centering
\setlength{\tabcolsep}{0pt}  
\fontsize{8}{10}\selectfont
\renewcommand{\arraystretch}{1.0}
\setlength{\fboxsep}{0pt}  
\colorbox{mycolor}{%
\begin{tabularx}{\columnwidth}{@{}X@{}}  
\toprule
\sloppy
The descriptions include a series of object motions related to free fall, collision, vibration, friction, fluid dynamics, projectile motion, and rotation. Based on these descriptions, four sets of videos are generated. Observing the motion trajectories of the objects in the generated videos, the videos are then ranked based on how well they adhere to real-world physical laws. Specifically, evaluators consider whether the object's motion trajectory, speed, and acceleration are reasonable and credible given the object's material, mass, volume, and the forces acting upon it, and whether such motion is likely to occur in the real world. Videos are ranked from best to worst using numerical labels, for example, ``2 $>$ 1 $>$ 3 $>$ 4,'' with the most physically plausible listed first. If the ranking cannot be determined between videos, use an equal sign, such as ``2 $>$ 1 = 3 $>$ 4.'' \par
Please do not focus on:\par
(1) The details of human faces or bodies in the video.\par
(2) The video style.\par
(3) Camera movements.\par
\\  
\bottomrule
\end{tabularx}%
}
\caption{\textbf{The requirements for manual evaluation.}}
\label{tab:manual}
\end{table}

\subsection{Correlation Coefficient Calculation}
When calculating the Kendall and Spearman correlation coefficients between the manual ranking results and the automated ranking results from our frame prediction model, we assign numerical values of 1, 2, 3, and 4 to the ranking order $A > B > C > D$. If two models are ranked equally in the manual evaluation, we take the average of their positions as the ranking value for both models. However, if more than two models are ranked equally, to avoid generating NaN in the calculations, we skip that particular sample.

\section{Training Details}
\label{sec:training}

Our training is divided into two stages. All of our training is conducted using 64 L20 GPUs.

\subsection{Stage One}
In the first stage, we begin by training the Latent Adapter and 3D U-Net in the Latent Flow Diffusion Module from scratch using the filtered OpenVid dataset, with a learning rate of 5e-5 and a batch size of 4, for a total of 100k steps. We then fine-tune the Latent Adapter and 3D U-Net using the Motion Data, with a reduced learning rate of 1e-5, keeping the batch size the same, and training for an additional 30k steps.

\subsection{Stage Two}
In the second stage, we train the Flow Adapter and 3D U-Net in the Latent Video Diffusion Module, using pre-trained weights for the U-Net provided by DynamiCrafter \cite{dynamicrafter}. The training also uses Motion Data, with a learning rate of 1e-5, a batch size of 4, for a total of 30k steps.

\end{document}